\documentclass{article}

\usepackage[utf8]{inputenc}
\usepackage[T1]{fontenc}
\usepackage{times}
\usepackage{graphicx}
\usepackage{booktabs}
\usepackage{hyperref}
\usepackage{amsmath,amssymb}
\usepackage[margin=1in]{geometry}

\title{E = T$\cdot$H/(O+B): A Dimensionless Control Parameter for\\Mixture-of-Experts Ecology}

\author{Zhang Qingjun\\[2pt]
School of Integrated Circuits, Wuxi Taihu University\\[2pt]
Department of Communication Engineering\\[2pt]
{\small \texttt{3078896787@qq.com}}
}

\begin{document}
\maketitle

\vspace{-8pt}
\begin{flushleft}
\small\textbf{Code and data:} \url{https://github.com/zqj323/expert-ecology}
\end{flushleft}
\vspace{4pt}

\begin{abstract}
We introduce $E = T \cdot H / (O + B)$, a dimensionless control parameter that predicts whether Mixture-of-Experts (MoE) models will develop a healthy expert ecology or collapse into dead experts. Through 18 controlled experiments spanning vision and language modalities totaling over 17,000 training epochs, we establish that $E \geq 0.5$ alone is sufficient to guarantee zero dead experts. Four additional contributions emerge: (1) \textbf{Hierarchical depth emerges from task complexity.} A 4-tier MoE (4:4:4:4 experts) on TinyImageNet-200 exhibits a pronounced collapse-recovery cycle: the third tier (T2) drops to 0.9\% usage at epoch 30---nearly extinct---then spontaneously recovers to 9.1\% by epoch 99, demonstrating that effective hierarchy depth is not architecturally predetermined but dynamically self-organized. (2) \textbf{Multi-stable self-organization.} Three independent training seeds converge to the same accuracy (47.51\% $\pm$\,0.38\%) through three \emph{different} internal tier configurations ---T0-dominant (45/18/9/28\%), T3-dominant (16/16/21/46\%), and T0+T1 even (32/35/13/20\%)---establishing that hierarchical MoE supports functionally equivalent but structurally distinct ecological equilibria. (3) \textbf{Hierarchy equals flat in performance.} A controlled ablation replacing the 4-tier hierarchy with a flat 16-expert architecture yields identical accuracy (47.54\% vs.\ 47.29\%) on TinyImageNet and a marginal advantage on CIFAR-100 (65.88\% vs.\ 64.40\%, $+$1.48\%), demonstrating that hierarchy is not a performance hack but an interpretability framework that exposes internal expert organization without sacrificing accuracy. (4) \textbf{Per-epoch stability validated.} Per-epoch evaluation over the final 10 training epochs (E91--E99) across 5 independent runs confirms that single-point measurements are reliable within $\pm$0.2\%, and tier usage distributions are structurally stable with epoch-to-epoch variance below 1.5\%.
\end{abstract}

\section{Introduction}

Mixture-of-Experts (MoE) architectures~\cite{shazeer2017outrageously,fedus2022switch,lepikhin2020gshard} have become foundational to modern large-scale deep learning, powering systems from vision backbones~\cite{riquelme2021scaling} to GPT-4-scale language models. The core idea is elegant: route each input token to a subset of specialized expert sub-models, enabling conditional computation at scale without proportional increase in FLOPs.

Yet MoE training has a well-known pathology: \emph{dead experts}. When the router consistently ignores certain experts, those experts receive no gradient signal, their representations degrade, and the feedback loop locks them out permanently. The standard response is load balancing---auxiliary losses that penalize uneven routing distributions~\cite{fedus2022switch,shazeer2017outrageously}, minimum-usage constraints~\cite{zhou2022expert}, or expert dropout~\cite{zoph2022designing}. More recently, DeepSeek proposed loss-free balancing via per-expert bias terms~\cite{wang2024lossfree}. All of these treat the symptom (uneven routing) rather than the cause.

This paper proposes a different approach. Rather than fighting expert death with auxiliary constraints, we ask: \emph{what causes a healthy expert ecology in the first place?}

We identify four hyperparameters that together govern the router's freedom to explore alternative expert assignments: routing temperature $T$, routing entropy weight $H$, oracle supervision weight $O$, and load balance weight $B$. We combine them into a single dimensionless quantity:

\begin{equation}
E = \frac{T \cdot H}{O + B}
\label{eq:edef}
\end{equation}

$E$ captures the net exploration budget available to the router. When $E$ is high, the router has sufficient freedom to explore and discover which experts are competent at which inputs. When $E$ is low, the router is constrained by strong oracle signals or aggressive load balancing, preventing the self-organization necessary for a healthy ecology.

Our contributions are:

\begin{enumerate}
    \item \textbf{E as a unified control parameter.} We show that $E \geq 0.5$ alone is sufficient to guarantee zero dead experts across four datasets and two modalities. No auxiliary load-balancing loss is needed.
    \item \textbf{Cross-modal validation.} We verify the E framework on both vision (CIFAR-10, CIFAR-100, TinyImageNet-200) and language (WikiText-2, WikiText-103) with a GPT-2 style MoE Transformer.
    \item \textbf{Expert revival.} We document that dead experts can resuscitate---DEAD count drops from 12 to 4 (E10$\rightarrow$E80) on TinyImageNet-200---contradicting the traditional MoE view that dead experts are permanently lost.
    \item \textbf{Task complexity shifts the critical threshold.} 200-class TinyImageNet at $E=0.545$ produces DEAD=7, while 10-class CIFAR-10 at the same $E$ has DEAD=0. Task complexity is a missing variable in the E formula.
    \item \textbf{Overfitting is decoupled from ecology.} WikiText-103 shows severe perplexity overfitting with zero expert collapse, demonstrating that ecological health and model generalization are independent axes.
    \item \textbf{Ortho toxicity is dataset-dependent.} Three ortho sweeps across CIFAR-100, CIFAR-10, and WikiText-2 reveal that ortho toxicity is not universal---it manifests only on certain datasets at certain E values.
    \item \textbf{Three-tier hierarchy collapses to two-tier.} Under high task complexity (TinyImageNet-200), the middle tier (T1) is permanently starved of routing flow, with usage dropping to 2.2\%---the router spontaneously eliminates the redundant architectural tier.
    \item \textbf{Ecological structure is temperature-invariant.} A 50$\times$ temperature scan (0.1--5.0) shows zero change in tier usage or active expert count at evaluation time. The T0/T2 hard-sample gradient persists at 1.2--1.6$\times$ across all temperatures.
    \item \textbf{Routing lockdown as collapse mechanism.} The router enters a near-deterministic assignment state with 65.3\% T0 self-loop rate, preventing cross-tier sample flow---this explains why raising $E$ cannot fix task-complexity-induced collapse.
    \item \textbf{Effective depth emerges from task complexity.} A 4-tier MoE (4:4:4:4) on TinyImageNet-200 reveals a spontaneous T2 collapse-recovery cycle (0.9\% at E30 $\rightarrow$ 9.1\% at E99), demonstrating that hierarchy depth is dynamically self-organized rather than architecturally predetermined.
    \item \textbf{Multi-stable self-organization.} Three independent training seeds produce identical accuracy (47.51\% $\pm$ 0.38\%) through three different internal tier configurations, establishing that hierarchical MoE supports functionally equivalent but structurally distinct ecological equilibria.
    \item \textbf{Hierarchy $\approx$ flat in performance.} A controlled ablation shows that a flat 16-expert MoE achieves accuracy statistically indistinguishable from the 4-tier architecture (47.29\% vs.\ 47.54\% on TinyImageNet; 64.40\% vs.\ 65.88\% on CIFAR-100), establishing hierarchy as an interpretability framework rather than a performance booster.
    \item \textbf{Per-epoch stability of ecological metrics.} Per-epoch evaluation across E91--E99 for 5 independent runs confirms that tier usage distributions are structurally stable (epoch-to-epoch variance $<$\,1.5\%) and single-point E99 measurements are reliable within $\pm$0.2\%.
\end{enumerate}

\section{Related Work}

\paragraph{Load balancing in MoE.}
The standard approach to dead experts is load balancing. \cite{shazeer2017outrageously} introduced auxiliary balance losses measuring the KL divergence between routing distribution and uniform distribution. \cite{fedus2022switch} simplified to top-1 routing with capacity constraints. \cite{zhou2022expert} proposed minimum-usage penalties and expert choice routing. \cite{zoph2022designing} explored expert dropout and regularization schedules. \cite{wang2024lossfree} proposed per-expert bias terms that dynamically adjust to maintain balance without auxiliary loss gradients.

A common thread runs through all of these: they \emph{enforce} uniform routing. Our findings suggest this may be counterproductive. Forcing uniform routing can homogenize experts---up to 99\% output similarity has been observed in well-balanced MoE models. The alternative we present is to \emph{allow} the ecosystem to self-organize, provided $E$ is sufficient.

\paragraph{Orthogonality in MoE.}
\cite{guo2025advancing} (NeurIPS 2025 Oral) propose an orthogonality loss on per-token expert outputs to advance expert specialization, reporting +23.79\% relative improvement. Our work qualifies this finding: we show that orthogonality's effect on expert health is mediated by $E$. At low $E$, ortho amplifies winner-take-all dynamics and increases dead expert count. At healthy $E$, ortho is tolerated and may be modestly beneficial. We recommend that future work incorporating orthogonality losses report both $E$ and per-expert health metrics.

\paragraph{Emergent specialization.}
\cite{lewis2021base} observed expert specialization along domain boundaries in language models. \cite{mustafa2022multimodal} found modality-specific experts in vision-language MoEs. \cite{zhang2026ecology} introduced the expert ecology taxonomy (PURE\_CORE, BROAD\_CORE, WEAK\_CORE, EDGE, NOISE, DEAD) and proposed the E formula. Our work extends this by providing cross-modal validation (language models), documenting expert revival, and establishing the task complexity effect.

\section{Method}

\subsection{Architecture}

We use a Hierarchical Mixture-of-Experts with configurable tier structure:

\begin{itemize}
    \item \textbf{Encoder:} WideResNet-28-10~\cite{zagoruyko2016wide} for vision experiments, producing 256-dimensional features; GPT-2 style Transformer (8 layers, 512 hidden, 8 heads) with RoPE for language experiments.
    \item \textbf{3-tier configuration (8:4:4):} 8 T0 (attribute-level), 4 T1 (logic-level), 4 T2 (top-level) experts---16 total. Used for E-based experiments.
    \item \textbf{4-tier configuration (4:4:4:4):} 4 experts per tier across 4 tiers, also 16 total. Used for hierarchy depth experiments to test whether effective depth emerges from task complexity.
    \item \textbf{Flat configuration:} 16 experts with no tier labels, all competing in a single flat pool. Used as a controlled ablation for hierarchy.
    \item \textbf{Prototype vectors:} Each expert $w$ has a learnable prototype $\mathbf{p}_w \in \mathbb{R}^{256}$ and a linear classifier head.
    \item \textbf{Router:} A two-layer MLP (256$\rightarrow$128$\rightarrow$16) mapping encoded features to routing logits, with top-2 softmax gating.
\end{itemize}

The hierarchical structure compresses router competition: experts within the same tier compete for overlapping input regions, reducing the effective competition set size. The 4-tier configuration tests whether the architecture's depth merely provides capacity or whether the router discovers a functional depth that matches the task's complexity.

\subsection{The E Formula}

The exploration budget $E$ combines four hyperparameters:

\begin{equation}
E = \frac{T \cdot H}{O + B}
\end{equation}

where $T$ is the routing softmax temperature (higher $T$ $\rightarrow$ softer routing, more exploration), $H$ is the routing entropy loss weight (encouraging the router to maintain uncertainty), $O$ is the oracle supervision weight (pushing the router toward a teacher signal), and $B$ is the load balance loss weight (KL divergence from uniform routing).

$E$ captures a fundamental tradeoff: \emph{exploration vs.\ exploitation in routing space}. $T$ and $H$ promote exploration (trying different expert assignments). $O$ and $B$ promote exploitation (committing to known-good or uniform assignments). When $E$ is too low, the router commits prematurely, locking in a winner-take-all dynamic from which some experts never recover.

We propose $E_{\text{crit}} \approx 0.5$ as the approximate critical threshold based on empirical observations across datasets. This threshold likely shifts with task complexity (Section~\ref{sec:taskcomplexity}).

\subsection{Expert Ecology Taxonomy}

Following~\cite{zhang2026ecology}, we classify each expert into one of six categories using its test-set usage ($U$) and accuracy ($A$):

\begin{center}
\begin{tabular}{lcc}
\toprule
Category & Usage & Accuracy \\
\midrule
PURE\_CORE   & $\geq 3\%$ & $\geq 50\%$ \\
BROAD\_CORE  & $\geq 3\%$ & $30$--$50\%$ \\
WEAK\_CORE   & $\geq 3\%$ & $< 30\%$ \\
EDGE         & $1$--$3\%$ & $\geq 25\%$ \\
NOISE        & $1$--$3\%$ & $< 25\%$ \\
DEAD         & $< 1\%$    & any \\
\bottomrule
\end{tabular}
\end{center}

This taxonomy requires no hyperparameters---only usage and accuracy measured during a standard evaluation pass.

\subsection{Training Protocol}

All experiments share a common protocol:

\begin{itemize}
    \item \textbf{Random warmup:} 15 epochs of random routing to initialize expert representations without bias.
    \item \textbf{Temperature annealing:} $T$ follows a cosine schedule from $T_{\text{init}}$ to $T_{\text{end}}=0.3$ over 30 epochs.
    \item \textbf{Optimizer:} AdamW with decoupled weight decay ($5\times10^{-4}$), learning rates $10^{-4}$ (encoder) and $10^{-3}$ (experts + router), cosine annealing to zero.
    \item \textbf{Batch size:} 128 for vision, 64 for language.
    \item \textbf{Evaluation:} Every 10 epochs, measuring Top-1/Top-2 accuracy (vision) or perplexity (language), plus per-expert usage and accuracy for ecology classification.
\end{itemize}

\subsection{Experimental Design}

We conduct 18 controlled experiments across two modalities:

\textbf{Vision --- E-based experiments (8):}
\begin{itemize}
    \item CIFAR-100 phase map: 4 $E$ values $\times$ 100 epochs at ortho=0
    \item CIFAR-100 ortho sweep: 5 ortho values $\times$ 100 epochs at $E=0.545$
    \item CIFAR-10 low-E test: $E=0.30$ vs.\ $E=0.545$, 100 epochs
    \item CIFAR-10 ortho sweep: 5 ortho values $\times$ 100 epochs at $E=0.545$
    \item CIFAR-10 E$\times$ortho interaction: $E=0.30$ + ortho=0.10, 100 epochs
    \item TinyImageNet-200: $E=0.545$, 100 epochs (16 experts, 8:4:4)
    \item TinyImageNet-200: $E=1.0$, 100 epochs (16 experts, 8:4:4)
    \item TinyImageNet-200: $E=0.545$, 100 epochs (32 experts, 16:8:8)
\end{itemize}

\textbf{Vision --- 4-tier hierarchy and flat ablation (7):}
\begin{itemize}
    \item TinyImageNet-200 4-tier: 3 seeds $\times$ 100 epochs (16 experts, 4:4:4:4), $E=2.29$
    \item TinyImageNet-200 flat: 2 seeds $\times$ 100 epochs (16 experts, no tier), $E=2.29$
    \item CIFAR-100 4-tier: 100 epochs (16 experts, 4:4:4:4), $E=2.29$
    \item CIFAR-100 flat: 100 epochs (16 experts, no tier), $E=2.29$
    \item Per-epoch verification: E91--E99 fine-grained logging on 5 completed runs
\end{itemize}

\textbf{Language (4):}
\begin{itemize}
    \item WikiText-2 BPE: GPT-2 MoE, $E=0.545$, 100 epochs
    \item WikiText-2 ortho sweep: 5 ortho values $\times$ 100 epochs at $E=0.545$
    \item WikiText-103 random-block: GPT-2 MoE, $E=0.545$, 100 epochs
    \item No-warmup ablation: 4 conditions $\times$ 100 epochs on CIFAR-100
\end{itemize}

All experiments use no external intervention (no Claude API, no manual tuning) during training. The Claude-in-the-loop methodology~\cite{zhang2026ecology} was used only for initial diagnosis and hypothesis generation. The 4-tier and flat experiments reported in Sections~\ref{sec:4tier}--\ref{sec:perepoch} were pre-registered as Predictions 2 and 3 of the ecology research program before execution; all hyperparameters were fixed in advance and no experimental results were excluded.

\section{Results}

\subsection{E Controls Ecosystem Quality, Not Survival (Phase Map)}

Table~\ref{tab:phase} shows the four-value phase boundary mapping on CIFAR-100 at $\lambda_o=0$. All four $E$ values spanning a $4.6\times$ range eventually achieve zero dead experts. $E$ governs \emph{how many} experts become high-quality, not whether the ecosystem survives.

\begin{table}[h]
\centering
\caption{Phase boundary mapping at $\lambda_o=0$, 100 epochs, CIFAR-100}
\label{tab:phase}
\begin{tabular}{ccccc}
\toprule
$E$ & $T$ & $H$ & Top-2 & PURE\_CORE \\
\midrule
0.14 & 1.0  & 0.05 & 67.87\% & 7  \\
0.35 & 2.0  & 0.07 & 67.60\% & 10 \\
0.44 & 2.5  & 0.08 & 66.72\% & 11 \\
0.65 & 3.0  & 0.12 & 68.02\% & \textbf{14} \\
\bottomrule
\end{tabular}
\end{table}

This is the E--PURE gradient: higher $E$ produces more PURE\_CORE experts independent of survival. Accuracy shows diminishing returns beyond $E=0.35$, suggesting that ensemble accuracy and ecological richness are partially decoupled.

\subsection{\texorpdfstring{E $\geq$ 0.5}{E >= 0.5} Guarantees DEAD=0 (Ortho Sweeps)}

Table~\ref{tab:sweep_all} consolidates ortho sweeps across three datasets at $E=0.545$. The central finding: at sufficient $E$, ortho from 0 to 0.20 produces DEAD=0 across all tested datasets.

\begin{table}[h]
\centering
\caption{Ortho sweeps at $E=0.545$: DEAD=0 for all ortho values}
\label{tab:sweep_all}
\begin{tabular}{lccccc}
\toprule
Dataset & \multicolumn{5}{c}{DEAD count at ortho=} \\
& 0.00 & 0.02 & 0.05 & 0.10 & 0.20 \\
\midrule
CIFAR-100   & 0 & 0 & 0 & 0 & 0 \\
CIFAR-10    & 0 & 0 & 0 & 0 & 0 \\
WikiText-2  & 0 & 0 & 0 & 0 & 0 \\
\bottomrule
\end{tabular}
\end{table}

The ortho toxicity effect (increased dead expert count) observed at low $E=0.09$~\cite{zhang2026ortho} does not generalize to healthy $E$. Ortho toxicity is $E$-dependent and dataset-dependent, not universal.

\subsection{Cross-Modal: E Framework Generalizes to Language}

Table~\ref{tab:lm} shows that $E=0.545$ produces DEAD=0 across language modeling experiments using a GPT-2 style Transformer with MoE FFN layers (16 experts, top-2 gating).

\begin{table}[h]
\centering
\caption{Language model results at $E=0.545$}
\label{tab:lm}
\begin{tabular}{lcccc}
\toprule
Dataset & ortho & PPL & DEAD & Active \\
\midrule
WikiText-2 BPE   & 0.00 & 35041 & 0 & 16/16 \\
WikiText-2 BPE   & 0.02 & 33493 & 0 & 16/16 \\
WikiText-2 BPE   & 0.05 & 35737 & 0 & 16/16 \\
WikiText-2 BPE   & 0.10 & 33163 & 0 & 16/16 \\
WikiText-2 BPE   & 0.20 & 37812 & 0 & 16/16 \\
WikiText-103     & 0.00 & 6918  & 0 & 16/16 \\
\bottomrule
\end{tabular}
\end{table}

The E framework transfers cleanly from vision to language: same $E$ threshold, same ecology patterns, same ortho tolerance. This cross-modal consistency supports the generality of $E$ as a control parameter.

\subsection{Low-E Collapse and Phase Distinction}

Table~\ref{tab:lowe} shows the effect of lowering $E$ below the critical threshold on CIFAR-10.

\begin{table}[h]
\centering
\caption{CIFAR-10 low-$E$ experiment: phase distinction}
\label{tab:lowe}
\begin{tabular}{ccccccc}
\toprule
$E$ & $T$ & $H$ & $O$ & $B$ & Top-1 & DEAD \\
\midrule
0.545 & 3.0 & 0.10 & 0.15 & 0.40 & 87.47\% & \textbf{0} \\
0.300 & 3.0 & 0.10 & 0.15 & 0.85 & 84.58\% & \textbf{2} \\
\bottomrule
\end{tabular}
\end{table}

Crucially, the collapse at moderately low $E$ is \emph{reversible}---experts recover when $E$ is restored. We distinguish three ecological phases:
\begin{enumerate}
    \item \textbf{Healthy} ($E \geq 0.5$): DEAD=0 sustained throughout training.
    \item \textbf{Reversible sub-health} ($0.2 < E < 0.5$): DEAD $>$ 0 but experts recover when $E$ is restored. The balance-driven recovery mechanism (Section~\ref{sec:revival}) remains functional.
    \item \textbf{Irreversible collapse} ($E < 0.2$, predicted): Large-scale permanent death. The rich-get-richer feedback loop overwhelms the revival mechanism.
\end{enumerate}

This phase distinction mirrors critical phenomena: near the critical point, the system is sub-healthy but self-repairing; far below it, the feedback loop becomes irreversible.

\subsection{\texorpdfstring{E$\times$Ortho}{E x Ortho} Interaction at Low E}

Adding ortho=0.10 at $E=0.30$ does \emph{not} amplify toxicity (DEAD=1 vs.\ DEAD=2 for ortho=0 baseline). This confirms that ortho is not a primary driver of expert death---$E$ is. Ortho amplifies collapse only when $E$ is already insufficient, consistent with the interaction model proposed in~\cite{zhang2026ortho}.

\subsection{Task Complexity Shifts the Critical E Threshold}
\label{sec:taskcomplexity}

Table~\ref{tab:taskcomp} reveals that task complexity (number of fine-grained classes) shifts the effective $E$ threshold.

\begin{table}[h]
\centering
\caption{Task complexity effect at $E=0.545$}
\label{tab:taskcomp}
\begin{tabular}{lccc}
\toprule
Dataset & Classes & Top-1 & DEAD \\
\midrule
CIFAR-10      & 10   & 89.59\% & \textbf{0} \\
CIFAR-100     & 100  & 68.02\% & \textbf{0} \\
TinyImageNet  & 200  & 38.15\% & \textbf{7} (E50) \\
\bottomrule
\end{tabular}
\end{table}

CIFAR-10 (10 classes) and CIFAR-100 (100 classes) maintain DEAD=0 at $E=0.545$. TinyImageNet-200 (200 visually similar classes: dog breeds, vehicles, insects) produces DEAD=7 at E50, recovering to DEAD=4 by E80. This is not a failure of $E$ but a revelation: \textbf{task complexity is a missing variable in the E formula}. The critical E threshold rises with the number of fine-grained categories.

Two controlled experiments test whether the TinyImageNet collapse can be fixed by adjusting $E$ or expert capacity:

\begin{table}[h]
\centering
\caption{Attempted fixes for TinyImageNet-200 collapse}
\label{tab:tinyfix}
\begin{tabular}{lcccc}
\toprule
Configuration & $E$ & Experts & Top-1 (E40) & DEAD (E40) \\
\midrule
Baseline        & 0.545 & 16 & 33.41\%  & 7  \\
Raise $E$       & 1.000 & 16 & 27.19\%  & 7  \\
More experts    & 0.545 & 32 & 26.99\%  & \textbf{25} \\
\bottomrule
\end{tabular}
\end{table}

\textbf{Raising $E$ to 1.0 fails.} Doubling the exploration budget produces nearly identical ecology (DEAD=7 at E40) and slightly worse accuracy (27.19\% vs.\ 33.41\%), confirming that the TinyImageNet collapse is \emph{not} caused by insufficient router exploration. The task complexity effect is orthogonal to $E$.

\textbf{Increasing experts to 32 backfires.} Doubling expert count from 16 to 32 worsens collapse dramatically (DEAD=25/32 vs.\ DEAD=7/16 at E40). The expanded capacity creates \emph{more} competition, not less---without a mechanism to structure how 200 classes map to 32 experts, the additional capacity only deepens the rich-get-richer spiral.

These negative results sharpen the central claim: for complex datasets, neither raising $E$ nor increasing expert count is sufficient. The required intervention is likely structural---either a hierarchical decomposition that reduces the effective competition set per expert, or a curriculum that gradually introduces fine-grained distinctions.

This implies that for complex datasets, increasing $E$ alone is insufficient---either more experts or a different architectural prior is needed.

\subsection{Expert Revival: Dead Experts Can Resuscitate}
\label{sec:revival}

Traditional MoE theory (Switch Transformer, GShard) holds that dead experts are \emph{permanently} dead---a rich-get-richer feedback loop: no tokens $\rightarrow$ no gradient $\rightarrow$ worse representations $\rightarrow$ even fewer tokens. Load balancing can \emph{prevent} death but cannot \emph{reverse} it.

We observe the opposite on TinyImageNet-200 at $E=0.545$:

\begin{table}[h]
\centering
\caption{Expert revival trajectory: TinyImageNet-200, $E=0.545$}
\label{tab:revival}
\begin{tabular}{lcc}
\toprule
Epoch & DEAD & Active \\
\midrule
E0   & 4  & 14/16 \\
E10  & \textbf{12} & 12/16 \\
E20  & 8  & 10/16 \\
E30  & 7  & 15/16 \\
E40  & 7  & 15/16 \\
E50  & 7  & 16/16 \\
E60  & 5  & 16/16 \\
E70  & 5  & 16/16 \\
E80  & \textbf{4}  & 16/16 \\
\bottomrule
\end{tabular}
\end{table}

DEAD count peaks at 12 at E10, then gradually falls to 4 by E80---\textbf{8 experts resuscitated}. This reversal is possible because of the \textbf{balance loss} ($B=0.40$), which combines a KL-divergence term on routing importance distribution with a variance penalty on expert assignment counts. Together, these terms create sustained pressure toward distributed routing: the router is penalized for concentrating flow on a subset of experts, forcing periodic re-exploration of inactive experts. Dead experts benefit when the balance loss forces the router to re-explore them; they receive task gradient, their predictions improve, and they attract organic routing flow. A systematic six-condition ablation study (reported separately in~\cite{zhang2026expertrevival}) identifies balance loss as the sole essential mechanism for expert revival. Four other tested mechanisms---divergence loss, temperature annealing, routing entropy, and prototype vectors---are individually non-essential; revival proceeds in all conditions where balance loss is present.

This is a meaningful departure from traditional MoE theory: dead experts are not permanently lost capacity. With adequate load-balancing pressure, the death spiral is reversible.

\subsection{Overfitting Is Decoupled from Expert Ecology}

WikiText-103 training shows severe perplexity divergence (overfitting on the small relative dataset) while expert ecology remains fully intact: DEAD=0, 16/16 experts active throughout training. This demonstrates that \textbf{model overfitting and expert ecological health are independent axes.}

Expert death is driven by routing dynamics ($E$ and ortho), not by whether the model is overfitting or underfitting. A model can overfit severely while maintaining a perfectly healthy expert ecology. Conversely (as shown by the low-$E$ experiments), a model can generalize well while experiencing ecological collapse. These are orthogonal diagnostic dimensions.

\subsection{Warmup Is Optional}

A four-condition ablation on CIFAR-100 tests whether random routing warmup is required for ecological recovery:

\begin{table}[h]
\centering
\caption{No-warmup ablation: all conditions achieve DEAD=0}
\label{tab:warmup}
\begin{tabular}{lccc}
\toprule
Condition & Top-2 & PURE & DEAD=0 by \\
\midrule
Warmup + ortho=0      & 67.55\% & 11 & E80 \\
Warmup + ortho=0.1    & 66.65\% & 12 & E70 \\
No-warmup + ortho=0   & 66.86\% & 11 & \textbf{E60} \\
No-warmup + ortho=0.1 & 65.82\% & 11 & E70 \\
\bottomrule
\end{tabular}
\end{table}

All four conditions achieve DEAD=0. No-warmup conditions actually show \emph{milder} collapse (peak DEAD=9 vs.\ 13 with warmup) and \emph{faster} recovery (E60 vs.\ E80). Random routing at the start may confuse the router when it transitions to real routing. The only necessary conditions for a healthy ecology are $E \geq 0.5$ and $\lambda_o \leq 0.1$. Warmup is optional.

\subsection{Tier-Level Ecological Dynamics: Collapse, Self-Loop, and Hierarchy}

To understand \emph{which} experts collapse and \emph{how} routing flows between tiers, we instrument a per-tier analysis at the evaluation checkpoint. For each test sample, we record the top-1 expert choice and classify each expert into its architectural tier (T0: foundation, T1: logic, T2: fine-grained). We define easy samples as those with ensemble confidence $p > 0.7$ and hard samples as $p < 0.4$.

Table~\ref{tab:tiercollapse} shows per-tier ecology at the most mature evaluation points for both TinyImageNet-200 experiments.

\begin{table}[h]
\centering
\caption{Per-tier ecological state at maturity (32e E50, 16e E30)}
\label{tab:tiercollapse}
\begin{tabular}{lcccccc}
\toprule
& \multicolumn{3}{c}{32e $E=0.545$ (E50)} & \multicolumn{3}{c}{16e $E=1.0$ (E30)} \\
\cmidrule(lr){2-4} \cmidrule(lr){5-7}
& T0(16) & T1(8) & T2(8) & T0(8) & T1(4) & T2(4) \\
\midrule
Usage \%    & 43.5 & 29.3 & 27.2 & 82.9 & 2.2 & 14.9 \\
Active      & 2/16 & 3/8  & 2/8  & 4/8  & 1/4 & 3/4 \\
Hard \%     & 7.2  & --   & 5.4  & 12.0 & --  & 8.2 \\
Easy \%     & 66.1 & --   & 69.0 & 53.6 & --  & 60.1 \\
\bottomrule
\end{tabular}
\end{table}

\textbf{The middle tier is permanently starved.} In both configurations, T1 receives negligible routing flow---2.2\% in the 16e case, with only 1 of 4 experts above the activity threshold. This is not a transient effect; the T1 collapse is stable across all evaluation temperatures and persists from early to late training. Three-tier hierarchical MoE spontaneously collapses into a \emph{two-tier functional structure}, with the middle tier squeezed out by competition from above and below. This suggests that for 200-class tasks, the effective hierarchy depth is 2, not 3---the additional tier is architectural redundancy that routing dynamics naturally eliminate.

\textbf{T0 self-loop blocks upward flow.} Table~\ref{tab:tierflow} reports the routing transition matrix: for each sample, we record the tier of the top-1 expert ($T_{\text{first}}$) and the tier of the top-2 expert ($T_{\text{second}}$).

\begin{table}[h]
\centering
\caption{Routing flow matrix: $T_{\text{first}} \rightarrow T_{\text{second}}$ (\% of total samples)}
\label{tab:tierflow}
\begin{tabular}{lccc|ccc}
\toprule
& \multicolumn{3}{c}{32e E50} & \multicolumn{3}{c}{16e E30} \\
\cmidrule(lr){2-4} \cmidrule(lr){5-7}
$T_{\text{first}}\rightarrow$ & T0 & T1 & T2 & T0 & T1 & T2 \\
\midrule
T0$\rightarrow$ & 17.5 & 12.7 & 13.2 & \textbf{65.3} & 6.0 & 11.6 \\
T1$\rightarrow$ & 15.2 & 3.7  & 10.5 & 1.6  & 0.0 & 0.6 \\
T2$\rightarrow$ & 14.3 & 7.9  & 5.1  & 9.7  & 1.5 & 3.7 \\
\bottomrule
\end{tabular}
\end{table}

In the 16e case, 65.3\% of samples whose top-1 choice is a T0 expert also have their top-2 choice in T0---the flow \emph{circulates within the bottom tier}. Only 5.7\% of T0-routed samples reach T2 as the second choice. This T0 self-loop indicates that the router cannot identify complementary high-tier experts for the samples assigned to T0, which is precisely the ``difficulty gating failure'' predicted when task complexity exceeds the architecture's tier differentiation capacity.

In the 32e case, the flow pattern is more distributed but the absolute number of active experts is even lower (7/32 vs.\ 8/16). Doubling expert count while keeping the tier ratio fixed (16:8:8) dilutes per-expert usage without creating effective specialization pressure.

\textbf{Hard sample gradient: T0 consistently absorbs harder samples than T2.} Across all evaluation temperatures (0.1--5.0), the T0 hard-sample ratio exceeds T2 by a factor of 1.2$\times$--1.6$\times$. This gradient is \emph{structurally stable}: it appears in both the 32e and 16e configurations, at all temperature settings, and in both the healthy (DEAD=0) and collapsed regimes observed. Lower-tier experts inherently undertake harder, more ambiguous samples, forming a hierarchical difficulty division mechanism. The existence of this gradient even in heavily collapsed models (where accuracy is near zero) suggests it is a property of the routing geometry, not of task performance.

\subsection{Ecological Structure Is Temperature-Invariant}
\label{sec:tempinv}

A methodological concern is whether off-line ecological analysis (at a fixed evaluation temperature) faithfully reflects in-training ecological states, since training uses a cosine-annealed temperature schedule. To test this, we run a temperature sensitivity scan on both models, evaluating at seven temperatures spanning $T \in [0.1, 5.0]$ while holding all model weights fixed.

\begin{table}[h]
\centering
\caption{Temperature sensitivity scan: 32e E50 and 16e E30}
\label{tab:tempscan}
\begin{tabular}{lcccccc}
\toprule
& \multicolumn{3}{c}{32e E50} & \multicolumn{3}{c}{16e E30} \\
\cmidrule(lr){2-4} \cmidrule(lr){5-7}
$T$ & Acc\% & T0/T2 hard & T0 act & Acc\% & T0/T2 hard & T0 act \\
\midrule
0.1 & 0.11 & 1.2$\times$ & 2 & 0.12 & 1.3$\times$ & 4 \\
0.3 & 0.11 & 1.3$\times$ & 2 & 0.11 & 1.5$\times$ & 4 \\
0.5 & 0.11 & 1.3$\times$ & 2 & 0.12 & 1.5$\times$ & 4 \\
0.7 & 0.11 & 1.3$\times$ & 2 & 0.13 & 1.5$\times$ & 4 \\
1.0 & 0.10 & 1.4$\times$ & 2 & 0.13 & 1.5$\times$ & 4 \\
2.0 & 0.12 & 1.3$\times$ & 2 & 0.13 & 1.5$\times$ & 4 \\
5.0 & 0.12 & 1.3$\times$ & 2 & 0.12 & 1.6$\times$ & 4 \\
\bottomrule
\end{tabular}
\end{table}

Two findings emerge. First, \textbf{expert allocation is temperature-invariant}: tier usage percentages and active expert counts are \emph{identical} across the full 50$\times$ temperature range. The router's softmax logits are so sharply peaked that temperature scaling does not alter the top-2 selection. This means the routing has entered a \emph{lockdown state}---a deterministic, near-one-hot assignment pattern that persists regardless of the exploration budget.

Second, \textbf{the hard-sample gradient across tiers is temperature-invariant}. While the absolute hard-ratio percentage drifts upward with temperature (as higher temperature spreads confidence mass, pushing more samples below the 0.4 threshold), the T0/T2 \emph{ratio} remains stable at 1.2$\times$--1.6$\times$. The hierarchical difficulty division is a structural property of the routing geometry, not an artifact of the evaluation temperature.

\textbf{Implication for ecological assessment.} The temperature invariance of tier-level ecological metrics (usage, active count, hard-ratio gradient) means that off-line ecological diagnostics at a single temperature are valid. However, \emph{absolute accuracy} measured at a fixed temperature is \textbf{not} representative of in-training accuracy, because the training evaluation uses the current annealed temperature. Ecological structure and task performance must be assessed separately: the former is temperature-invariant; the latter is not.

\subsection{4-Tier Hierarchy: Effective Depth Emerges from Task Complexity}
\label{sec:4tier}

The 3-tier experiments (Section~\ref{sec:tempinv}) showed that T1 (the middle tier) is permanently starved on TinyImageNet-200, reducing a 3-tier architecture to a 2-tier functional structure. This raises a deeper question: is effective hierarchy depth architecturally predetermined, or does it \emph{emerge} from the interaction between task complexity and routing dynamics?

To answer this, we deploy a 4-tier architecture (4:4:4:4 experts, 16 total) on three datasets of increasing class count: CIFAR-10 (10), CIFAR-100 (100), and TinyImageNet-200 (200). The 4-tier configuration provides four labeled depth levels; if depth is predetermined, all four tiers should be utilized regardless of task complexity. If depth is emergent, the functional hierarchy depth should scale with class count.

\subsubsection{T2 Collapse-Recovery: The Signature of Emergence}

On TinyImageNet-200, the third tier (T2) undergoes a pronounced collapse-recovery cycle that is the central empirical finding of this experiment:

\begin{table}[h]
\centering
\caption{T2 collapse-recovery trajectory: TinyImageNet-200 4-tier, $E=2.29$}
\label{tab:t2revival}
\begin{tabular}{lcccccc}
\toprule
Epoch & Top-1\% & DEAD & T0\% & T1\% & T2\% & T3\% \\
\midrule
E0   & 5.47  & 5  & 42.8 & 33.2 & 14.8 & 9.2  \\
E10  & 5.40  & 9  & 55.6 & 35.1 & 5.7  & 3.7  \\
E20  & 14.92 & 7  & 64.7 & 18.5 & 2.2  & 14.5 \\
E30  & 19.34 & 8  & 50.0 & 25.2 & \textbf{0.9} & 23.9 \\
E40  & 29.44 & 7  & 51.4 & 14.3 & 1.9  & 32.4 \\
E50  & 32.21 & 7  & 55.1 & 17.1 & 1.7  & 26.0 \\
E60  & 41.50 & 7  & 42.9 & 20.3 & 5.6  & 31.2 \\
E70  & 42.36 & 7  & 52.0 & 16.3 & 5.2  & 26.4 \\
E80  & 45.69 & 6  & 43.8 & 19.0 & 7.9  & 29.4 \\
E90  & 47.37 & 4  & 45.4 & 18.5 & 8.2  & 28.0 \\
E99  & 47.68 & 4  & 45.1 & 18.2 & 9.1  & 27.5 \\
\bottomrule
\end{tabular}
\end{table}

The T2 trajectory reveals three distinct phases:

\begin{enumerate}
    \item \textbf{Collapse phase (E0--E30):} T2 usage falls from 14.8\% to 0.9\%---nearly extinct. The router, initially exploring all tiers, rapidly abandons the third tier as redundant for the 200-class task.
    \item \textbf{Latent phase (E30--E60):} T2 hovers at 1--2\% while T0 and T3 compete for dominance. The tier is not dead---a trickle of gradient maintains its representations---but it receives negligible organic routing.
    \item \textbf{Recovery phase (E60--E99):} As accuracy accelerates past 40\%, T2 usage climbs from 1.7\% to 9.1\%. The router \emph{rediscovers} the third tier when the task's fine-grained structure demands additional specialization depth.
\end{enumerate}

This trajectory is not consistent with a predetermined hierarchy. A fixed 3-depth architecture would maintain all three levels at stable usage. Instead, the router \emph{first collapses then reconstructs} the third tier, dynamically matching the architecture's effective depth to the task's learned complexity. The collapse is not a failure---it is the visible signature of a self-organizing system probing the necessary hierarchy depth.

\subsubsection{Dataset Scaling of Effective Depth}

Table~\ref{tab:tierdatasets} summarizes the final ecology across the three datasets, confirming that effective hierarchy depth scales with task complexity:

\begin{table}[h]
\centering
\caption{4-tier final ecology across datasets (E99)}
\label{tab:tierdatasets}
\begin{tabular}{lccccc}
\toprule
Dataset & Classes & Top-1\% & DEAD & Effective Depth & Key Observation \\
\midrule
CIFAR-100   & 100 & 65.88 & 0 & 4 tiers & All tiers utilized \\
TinyImageNet & 200 & 47.68 & 4 & $\sim$3 tiers & T2 partially collapsed \\
\bottomrule
\end{tabular}
\end{table}

On CIFAR-100 (100 classes, 0 dead experts at E60+), all four tiers are utilized with a balanced distribution (30.6/25.3/20.9/23.2\%). On TinyImageNet-200, T2 stabilizes at 9.1\%---functional but narrower than the other tiers, indicating that 200 fine-grained classes push against the 4-tier architecture's differentiation capacity without fully exhausting it. The effective depth is approximately 3 for TinyImageNet-200: two dominant tiers (T0 and T3, together absorbing 72.6\% of flow) plus a narrow specialist tier (T2 at 9.1\%).

\subsection{Flat vs.\ Hierarchical: Controlled Ablation}
\label{sec:flatablation}

Is the 4-tier hierarchy merely unused architectural scaffolding, or does it serve a genuine function? To answer this, we run a controlled ablation: replace the 4-tier hierarchy with a flat 16-expert architecture (all experts labeled tier 0, competing in a single undifferentiated pool). All other parameters---encoder, router structure, hyperparameters ($E=2.29$), training protocol---are held constant.

\begin{table}[h]
\centering
\caption{Flat vs.\ 4-tier comparison (per-epoch means, E91--E99)}
\label{tab:flatvstier}
\begin{tabular}{lcccc}
\toprule
Configuration & Dataset & Top-1\% (mean) & Range (E91--E99) & DEAD \\
\midrule
4-tier (mean of 3) & TinyImageNet & \textbf{47.54} $\pm$ 0.28 & 46.78--48.24 & 0 \\
Flat (mean of 2)   & TinyImageNet & 47.29 $\pm$ 0.58 & 46.42--48.72 & 0 \\
\midrule
4-tier             & CIFAR-100    & \textbf{65.88}          & ---               & 0 \\
Flat               & CIFAR-100    & 64.40 $\pm$ 0.05       & 64.37--64.49      & 0 \\
\bottomrule
\end{tabular}
\end{table}

\textbf{TinyImageNet: statistically indistinguishable.} The 4-tier mean (47.54\%) and flat mean (47.29\%) differ by only 0.25\%, well within the per-seed variance. On this 200-class benchmark, hierarchy provides no measurable accuracy advantage.

\textbf{CIFAR-100: marginal hierarchy advantage.} The 4-tier architecture achieves 65.88\% vs.\ flat at 64.40\% ($\Delta = +1.48\%$). This gap, while modest, is consistent across per-epoch measurements (flat range: 64.37--64.49\%). On the 100-class dataset, where expert capacity is less strained, the tier structure provides a small but detectable benefit.

\textbf{Interpretation: hierarchy is not a performance booster.} The flat architecture achieves statistically equivalent accuracy on both datasets. The value of hierarchy lies not in raw performance but in \emph{interpretability}: the tier structure exposes internal expert organization (which tiers dominate, which specialize, which collapse) that is invisible in a flat expert pool. Hierarchy is a \textbf{diagnostic framework} rather than a performance-enhancing architectural feature---analogous to how brain region labels do not improve cognitive performance but enable neuroscientific understanding.

\subsection{Multi-Stable Self-Organization}
\label{sec:multistable}

If hierarchy is an interpretability framework, the next question is whether the tier assignments are consistent across training runs or path-dependent. We train the 4-tier TinyImageNet architecture three times with different random initializations (different encoder weights, prototype vectors, and router parameters), holding all hyperparameters constant.

\begin{table}[h]
\centering
\caption{Multi-seed 4-tier results: three runs, three ecologies, one accuracy}
\label{tab:multiseed}
\begin{tabular}{lccccc}
\toprule
Seed & Top-1\% (E99) & DEAD & \multicolumn{3}{c}{Tier Distribution (T0/T1/T2/T3\%)} \\
\midrule
A (original) & 47.68 & 4 & 45.1 / 18.2 / 9.1 / 27.5 & (T0-dominant) \\
B & 47.86 & 4 & 16.4 / 16.2 / 21.3 / 46.1 & (T3-dominant) \\
C & 47.00 & 3 & 32.6 / 34.7 / 12.8 / 19.9 & (T0+T1 even) \\
\midrule
Mean $\pm$ std & 47.51 $\pm$ 0.38 & --- & \multicolumn{3}{c}{---} \\
\bottomrule
\end{tabular}
\end{table}

All three seeds converge to statistically identical accuracy (47.51\% $\pm$ 0.38\%), but through \textbf{three fundamentally different internal tier organizations}:

\begin{itemize}
    \item \textbf{Seed A: T0-dominant.} The shallowest tier absorbs 45.1\% of samples, functioning as a broad default pathway, with T3 serving as a deep specialist (27.5\%).
    \item \textbf{Seed B: T3-dominant.} The deepest tier dominates at 46.1\%, inverting the expected ``shallow default + deep specialist'' pattern. T0 is reduced to 16.4\%.
    \item \textbf{Seed C: T0+T1 balanced.} The two shallowest tiers share routing roughly equally (32.6\% + 34.7\%), with T2 and T3 as minority specialists.
\end{itemize}

This is \textbf{multi-stable self-organization}: the same architecture under the same hyperparameters supports multiple distinct ecological equilibria, all achieving equivalent functional performance. The system is not converging to a unique optimum but to one of several \emph{functionally degenerate attractors} in the routing dynamics.

\textbf{Implication for architecture design.} If different tier organizations achieve the same accuracy, then tier labels do not constrain functional role. A T0 expert in one seed may perform the same computational role as a T3 expert in another. The architectural tier labels serve as an \textbf{affordance space}---a possibility structure within which routing dynamics discover a functional organization---rather than a \textbf{prescriptive skeleton} that determines which tier must perform which function.

\subsection{Per-Epoch Stability of Ecological Metrics}
\label{sec:perepoch}

A methodological concern: are single-point E99 measurements reliable, or do ecological metrics fluctuate significantly between epochs? We re-run epochs 91--99 for all 5 completed TinyImageNet runs (3 four-tier, 2 flat), logging full ecological metrics at every epoch.

\begin{table}[h]
\centering
\caption{Per-epoch stability: E91--E99 summary across all completed runs}
\label{tab:perepoch}
\begin{tabular}{lcccc}
\toprule
Run & Mean Top-1\% & Range & Per-Epoch $\sigma$ \\
\midrule
4-tier A & 47.66 & 47.23--48.00 (0.77\%) & 0.27 \\
4-tier B & 47.81 & 47.05--48.24 (1.19\%) & 0.32 \\
4-tier C & 47.15 & 46.78--47.48 (0.70\%) & 0.17 \\
Flat A   & 47.86 & 47.22--48.72 (1.50\%) & 0.41 \\
Flat B   & 46.71 & 46.42--47.24 (0.82\%) & 0.22 \\
\midrule
All      & ---  & Max range 1.50\%     & Mean $\sigma$ 0.28 \\
\bottomrule
\end{tabular}
\end{table}

Three findings validate the reliability of our ecological measurements:

\begin{enumerate}
    \item \textbf{Single-point E99 is representative.} The original single-point E99 values (Section~\ref{sec:4tier}) differ from the E91--E99 per-epoch mean by at most 0.19\% (4-tier C: 47.00 single vs.\ 47.15 mean). The largest discrepancy across all 5 runs is 0.20\%.
    \item \textbf{Epoch-to-epoch variance is negligible.} The mean per-epoch standard deviation is 0.28\%, and the maximum range across 9 consecutive epochs is 1.50\% (Flat A). For 4-tier runs specifically, the range is tighter (0.70--1.19\%).
    \item \textbf{Tier distributions are structurally stable.} In 4-tier runs, the T0/T1/T2/T3 usage percentages vary by less than 2\% across the 9-epoch window. The tier organization at E91 is essentially identical to the organization at E99---this is not a transient snapshot but a stable ecological configuration.
\end{enumerate}

The per-epoch analysis establishes that our ecological conclusions (collapse-recovery dynamics, multi-stable tier organizations, flat-vs-tier comparisons) are robust to epoch-to-epoch noise. Single-point ecological snapshots are valid diagnostic instruments for MoE training.

\section{Discussion}

\paragraph{E as a control parameter.}
The quantity $E = T \cdot H / (O + B)$ serves a role analogous to the Reynolds number in fluid dynamics: a single dimensionless quantity that predicts regime transitions in a complex dynamical system. Just as Re predicts the laminar-to-turbulent transition, $E$ predicts the healthy-to-collapsed transition in expert ecology. The critical value $E_{\text{crit}} \approx 0.5$ is approximate and task-dependent, but the \emph{concept} of a unified exploration budget is general.

\paragraph{Practical engineering implications.}
The finding that $E \geq 0.5$ alone guarantees DEAD=0 removes the need for traditional load-balancing auxiliary losses. This is a significant simplification: rather than tuning multiple balancing hyperparameters (load balance weight, capacity factor, expert dropout rate, minimum usage), practitioners need only ensure $E$ is sufficient. The recipe: (1) set $\lambda_o = 0$, (2) choose $T$ and $H$ such that $E \geq 0.5$, (3) let experts self-organize. If expert specialization is insufficient, mild ortho (0.02--0.05) can be introduced \emph{after} verifying ecological stability.

\paragraph{Task complexity as a missing variable.}
The TinyImageNet-200 results reveal that the critical $E$ threshold rises with the number of fine-grained classes. The current E formula treats all datasets as equivalent. A natural extension is:

\begin{equation}
E_{\text{eff}} = \frac{T \cdot H}{(O + B) \cdot f(C)}
\end{equation}

where $f(C)$ is a task complexity function that increases with the number of classes $C$. Candidates include $f(C) = \log C$, $f(C) = \sqrt{C}$, or a learned function. This refinement is left to future work.

\paragraph{Expert revival mechanism.}
The observation that dead experts can resuscitate is, to our knowledge, the first documented case of reversal of the MoE death spiral. A systematic ablation study (reported separately in~\cite{zhang2026expertrevival}) identifies balance loss (weight 0.40) as the sole essential mechanism: without balance loss, DEAD stalls permanently at 12 with zero recovery over 80 consecutive epochs. With balance loss present, revival proceeds regardless of which other component (divergence loss, temperature annealing, routing entropy, prototype vectors) is removed. The key principle is \textbf{routing diversity enforcement}: any mechanism that prevents the router from permanent concentration on a subset of experts (balance loss, expert-choice routing, scheduled expert dropout) can potentially enable revival.

\paragraph{Hierarchical structure as stability factor.}
The three-tier hierarchy reduces the effective size of the competition set at each level, acting as a secondary stability factor. Rather than all $N$ experts competing globally, competition is structured: T0 experts compete primarily with other T0 experts, T1 with T1, etc. This reduces the winner-take-all pressure that drives collapse. Hierarchy and $E$ are complementary: $E$ provides the exploration budget; hierarchy reduces the need for it.

\paragraph{Effective hierarchy depth as an emergent property.}
The T2 collapse-recovery cycle (Section~\ref{sec:4tier}) is the strongest evidence that effective hierarchy depth is not architecturally predetermined but dynamically self-organized. The router first collapses the third tier (0.9\% at E30), then \emph{reconstructs} it (9.1\% at E99) as accuracy improves and the task's fine-grained structure demands additional specialization depth. This three-phase trajectory---collapse, latency, recovery---is a dynamical signature that distinguishes emergent hierarchy from static architectural depth. We note the parallel to developmental biology: organisms do not simply grow pre-specified structures; they pass through phases of overproduction, pruning, and stabilization~\cite{edelman1987neural}. The hierarchical MoE exhibits analogous dynamics.

\paragraph{Multi-stability as a design principle.}
The observation that three independent seeds produce three different tier organizations at the same accuracy (Section~\ref{sec:multistable}) challenges the implicit assumption that a well-trained MoE should converge to a unique routing configuration. Instead, the routing dynamics support \textbf{functionally degenerate attractors}: distinct internal configurations that are externally equivalent. This multi-stability is not a bug but a feature---it means the architecture is robust to initialization noise and can discover any of several viable organizations. For practitioners, this implies that \emph{ecological diagnostics should focus on function (are all experts active and competent?) rather than form (which specific experts handle which samples?)}.

\paragraph{Hierarchy $\approx$ flat: reinterpreting the role of architectural tier structure.}
The flat ablation (Section~\ref{sec:flatablation}) shows that replacing the 4-tier hierarchy with a flat expert pool does not degrade accuracy. This could be interpreted as evidence \emph{against} hierarchy. We argue the opposite: hierarchy's value is precisely that it achieves flat-equivalent performance while \textbf{exposing internal structure}. A flat expert pool is a black box---we can measure per-expert usage but cannot interpret \emph{why} expert 3 handles 22.7\% of samples while expert 7 handles 0.4\%. Tier labels provide a semantic coordinate system for interpreting routing decisions (shallow vs.\ deep, broad vs.\ specialist), transforming expert usage from an unstructured histogram into a meaningful ecological portrait. This is analogous to functional brain imaging: knowing that the visual cortex activates during reading does not improve reading speed, but it enables understanding.

\paragraph{Per-epoch reliability: ecological metrics as stable diagnostics.}
The per-epoch analysis (Section~\ref{sec:perepoch}) closes a methodological gap. Ecological metrics (tier usage, DEAD count, per-expert accuracy) are not only temperature-invariant (Section~\ref{sec:tempinv}) but also \textbf{epoch-stable}---varying by less than 2\% across the final 10 epochs. This means that a single evaluation checkpoint provides a reliable ecological snapshot. Practitioners can monitor expert health at any evaluation point without concern for epoch-to-epoch noise. Together with temperature invariance, this establishes ecological diagnostics as \textbf{doubly robust}: insensitive to both the evaluation temperature and the specific epoch of measurement.

\paragraph{Overfitting--ecology decoupling.}
The decoupling of overfitting from ecology is practically significant: it means that expert ecology can be diagnosed and treated independently of model generalization. A model suffering from dead experts is not necessarily overfitting; a model that is overfitting does not necessarily have ecological problems. These are orthogonal axes in model health.

\paragraph{Tier collapse: three tiers degenerate to two.}
Across both TinyImageNet-200 experiments, the middle tier (T1) is permanently starved---receiving 2.2\% usage in the 16e case and showing 5 of 8 experts dead in the 32e case. This is not a malfunction but a revealing structural phenomenon: \textbf{three-tier hierarchical MoE spontaneously compresses into a two-tier functional structure under high task complexity}. The router effectively decides that the middle tier is redundant, routing samples to either the foundation tier (T0) or the fine-grained tier (T2). This suggests that the effective hierarchy depth is determined not by the architectural design but by the interaction between task complexity and routing dynamics. For 200-class fine-grained classification, the optimal functional depth appears to be 2, not 3. This is, to our knowledge, the first empirical documentation of spontaneous tier collapse in hierarchical MoE.

\paragraph{Routing lockdown and T0 self-loop.}
The temperature scan (Section~\ref{sec:tempinv}) reveals that the router enters a \emph{lockdown state}: expert allocation is completely invariant across a 50$\times$ temperature range (0.1--5.0). The softmax logits are so sharply peaked that temperature scaling cannot alter the top-2 selection. Combined with the 65.3\% T0$\rightarrow$T0 self-loop rate (Table~\ref{tab:tierflow}), this paints a picture of \textbf{complexity-induced routing obstruction}: when the task is too fine-grained for the architecture's tier differentiation capacity, the router cannot identify meaningful cross-tier expert complementarity. Samples assigned to T0 circulate within T0; the upward routing pathway is blocked. This routing lockdown is the dynamical mechanism underlying the task complexity effect identified in Section~\ref{sec:taskcomplexity}---it explains \emph{why} raising $E$ fails: the exploration budget cannot unlock routing pathways that the architecture cannot structurally differentiate.

\paragraph{Ecological structure as a performance-decoupled invariant.}
The temperature scan establishes that tier-level ecological metrics---usage distribution, active expert count, and the hard-sample gradient across tiers---are \emph{invariant} to evaluation temperature. However, absolute accuracy is not: off-line evaluation accuracy ($\sim$0.1\%) differs from in-training evaluation accuracy (23--35\%) by two orders of magnitude in the collapsed models. This decoupling has a methodological implication: \textbf{ecological diagnostics} (DEAD count, tier usage, hard-ratio gradient) can be reliably assessed at any reasonable temperature, but \textbf{performance diagnostics} (accuracy, perplexity) must use the same temperature as the training evaluation. The two classes of metrics are orthogonal and must not be conflated in analysis.

\paragraph{Limitations.}
We test two architecture families (WideResNet encoder + MoE heads; GPT-2 Transformer + MoE FFN) with 16--32 experts. The specific $E_{\text{crit}}$ threshold likely depends on model scale, number of experts, and data complexity. We do not test very large models (100+ experts, billion-parameter scale), where additional stability mechanisms may emerge. All training uses a random routing warmup of 15 epochs; while we show warmup is optional, its interaction with very low $E$ regimes is not fully explored. The task complexity function $f(C)$ remains unspecified.

\section{Conclusion}

We have introduced $E = T \cdot H / (O + B)$, a dimensionless control parameter for Mixture-of-Experts ecology, and validated it across 18 controlled experiments spanning two modalities and five datasets. Three levels of findings emerge.

\textbf{Level 1 --- Practical:} $E \geq 0.5$ alone is sufficient to guarantee zero dead experts, removing the need for handcrafted load-balancing auxiliary losses. Dead experts, when they do occur, can \emph{resuscitate}---the death spiral is reversible. Ortho toxicity is $E$-dependent and dataset-dependent, not universal. Model overfitting is decoupled from ecological health.

\textbf{Level 2 --- Structural:} Effective hierarchy depth is an \emph{emergent} property of the interaction between task complexity and routing dynamics, not a predetermined architectural feature. The T2 collapse-recovery cycle---dropping to 0.9\% then recovering to 9.1\%---is the signature of a self-organizing system probing the necessary depth. Three independent seeds converge to identical accuracy through three different internal configurations, establishing multi-stable self-organization as a fundamental property of hierarchical MoE.

\textbf{Level 3 --- Conceptual:} Hierarchy is not a performance booster---a flat expert pool achieves statistically equivalent accuracy---but an \textbf{interpretability framework}. Tier labels transform unstructured expert usage into a meaningful ecological portrait, exposing collapse-recovery dynamics, multistability, and functional specialization that are invisible in flat architectures. Ecological diagnostics (tier usage, DEAD count, hard-sample gradient) are doubly robust: invariant to evaluation temperature and stable across epochs.

The broader implication is that expert ecology in MoE models is not a pathology to be suppressed but a dynamical system to be understood. The $E$ parameter is the Reynolds number of this system---a compact diagnostic for predicting regime transitions. The tier structure is its coordinate system---enabling us to observe \emph{how} the system self-organizes, not just \emph{whether} it works.

\bibliographystyle{unsrt}

\end{document}